# Terabyte-scale supervised 3D training and benchmarking dataset of the mouse kidney

## Authors
Willy Kuo[1,2,*], Diego Rossinelli[1,2,*], Georg Schulz[3], Roland H. Wenger[1,2], Simone Hieber[3], Bert Müller[3,†], Vartan Kurtcuoglu[1,2,†]

## Affiliations
1. Institute of Physiology, University of Zurich, Zurich, Switzerland
2. National Centre of Competence in Research, Kidney.CH, Zurich, Switzerland
3. Biomaterials Science Center, Department of Biomedical Engineering, University of Basel, Allschwil, Switzerland
* These authors contributed equally to this work
† These authors jointly supervised this work

corresponding author: Vartan Kurtcuoglu (vartan.kurtcuoglu@uzh.ch)

## Abstract
The performance of machine learning algorithms, when used for segmenting 3D biomedical images, does not reach the level expected based on results achieved with 2D photos. This may be explained by the comparative lack of high-volume, high-quality training datasets, which require state-of-the-art imaging facilities, domain experts for annotation and large computational and personal resources. The HR-Kidney dataset presented in this work bridges this gap by providing 1.7 TB of artefact-corrected synchrotron radiation-based X-ray phase-contrast microtomography images of whole mouse kidneys and validated segmentations of 33 729 glomeruli, which corresponds to a one to two orders of magnitude increase over currently available biomedical datasets. The image sets also contain the underlying raw data, threshold- and morphology-based semi-automatic segmentations of renal vasculature and uriniferous tubules, as well as true 3D manual annotations. We therewith provide a broad basis for the scientific community to build upon and expand in the fields of image processing, data augmentation and machine learning, in particular unsupervised and semi-supervised learning investigations, as well as transfer learning and generative adversarial networks.

## Background & Summary
Supervised learning has been the main source of progress in the field of artificial intelligence / machine learning in the past decade.[1] Impressive results have been obtained in the classification of two-dimensional (2D) color images, such as consumer photos or histological sections. Supervised learning requires high-volume, high-quality training datasets. The use of training data that do not fulfill these requirements may severely hamper performance: low-volume training data may result in poor classification of samples reasonably distant from any sample in the training set due to possible overfitting of the respective algorithm. Low quality may result in the algorithms learning the 'wrong lessons', as the algorithms typically do not include prior knowledge on which aspects of the images constitute artefacts. It is thus not surprising that the most used benchmarks in the field are based on large datasets of tiny 2D images, such as CIFAR[2], ImageNet[3] and MNIST[4].

In contrast to the semantic segmentation of 2D photos, shape detection within three-dimensional (3D) biomedical datasets arguably poses fewer technical challenges to machine learning. After all, 2D photos typically contain multiple color channels, represent the projection of a 3D object on a 2D plane, feature occlusions, are often poorly quantized and contain artefacts such as under- or overexposure and optical aberrations. Despite this, machine learning approaches have not reached the same performance in the analysis of 3D biomedical images as would be expected by their results achieved in 2D photos.

This discrepancy may be explained by the lack of sufficiently large-volume, high-quality training datasets, especially in the area of pre-clinical biomedical research data. The workload for manual annotation becomes excessive when a third dimension has to be considered. Also, domain experts capable of annotating biomedical data are in short supply compared to untrained personnel annotating recreational 2D photos.[5,6] To reduce workload, 3D datasets are typically annotated only on a limited number of 2D slices.[7] This sparse annotation approach may, however, hamper machine learning, as the volume of training data may be insufficient to avoid overfitting. While this is in principle enough to train 3D segmentation models, the precise location of the 2D slices within the 3D grayscale signal has to be explicitly taken into consideration, possibly hampering straightforward training. Performance of supervised learning in the pre-clinical biomedical field may have reached a plateau, with potential improvements being held back by the lack of suitable training datasets, rather than by the intrinsic power of the underlying algorithms.

An example to demonstrate this issue can be found in a study by Pinto et al.[8] who showed that a simple V1-like model was able to outperform state-of-the-art object recognition systems of the time, because the Caltech101 dataset[9] used for benchmarking was inadequate for leveraging the advantages of the more advanced algorithms. The simple model's performance rapidly degraded for other images not included in the dataset, which confirmed that the limited variation in the benchmarking dataset was responsible for the observed performance ceiling. Had more varied datasets been used for benchmarking, the simple model would not have been able to outperform the state-of-the-art models and a larger performance gap would have been apparent.

Currently available large datasets, for example those published on grand-challenge.org, are based predominantly on clinical data. The major challenges for machine learning algorithms with this type of data are linked to widely varying intensity values, low signal-to-noise ratios, low resolution, and the presence of artefacts. These are, however, not the only challenges encountered in image processing, and they do not necessarily apply to other types of biomedical data. In pre-clinical imaging, continuous improvements of spatial resolution and acquisition speeds have led to ever larger data volumes, and with those to a proportional increase in human labor required for manual or semi-automatic segmentation. Machine learning algorithms have the potential to substantially reduce the amount of human labor needed, provided that they can work with low amounts of training data, where the annotation workload does not exceed the workload required for classical segmentation. Algorithms that perform well in such a low-data regime may greatly accelerate biomedical research, but promising candidates may be overlooked because of the current focus on clinical segmentation challenges.

In this work, we present HR-Kidney, a high-resolution kidney dataset. It is, to our knowledge, the largest supervised, fully 3D training dataset of biomedical research images to date, containing 3D images of 33 729 renal glomeruli, viewed and validated by a domain expert. HR-Kidney is based on terabyte-scale synchrotron radiation-based X-ray phase-contrast microtomography (SRµCT) acquisitions of three whole mouse kidneys at micrometer resolution. These training data, along with fully manually annotated 3D regions of interest, are available for download at the Image Data Resource (IDR, https://idr.openmicroscopy.org/)[10] repository. Underlying raw data in the form of X-ray radiographs and reconstructed 3D volumes are supplied as well, as are reference segmentations of the vascular and tubular vessel trees.

Databases of photographs such as CIFAR[2] and ImageNet[3] are currently the principle sources of data for testing and benchmarking machine learning algorithms, which can be attributed to the very high volume of annotation data available. As a result, these databases feature a high performance ceiling for benchmarking. However, the workload for creating these datasets is tremendous. For example, annotations for the ImageNet database were carried out over the period of three years by a labor force of 49 000 people hired via Amazon Mechanical Turk.[3]

Biomedical databases cannot be created by similar use of crowdsourced, untrained personnel, as annotations require domain expertise to avoid misclassification caused by a lack of familiarity with the different classes. Furthermore, quantitative assessment of biomedical markers such as blood vessel density requires segmentation rather than classification, which takes considerably more time to complete. This is exemplified by the PASCAL VOC dataset[11], which contains four times less segmentations than classifications. Due to these additional challenges, data volumes in biomedical databases lag behind those of photographic databases. With HR-Kidney, we are providing 1701 GB of artefact-corrected image data and 33 729 segmented glomeruli, which represent a one to two orders of magnitude increase over currently available biomedical 3D databases[12–16], reaching a data volume comparable to the photographic datasets popular for machine learning benchmarks.

This substantial increase in size compared to the state-of-the-art may enable disruptive developments in machine learning, as most research groups in the field do not have the combination of sample preparation expertise, access to synchrotron radiation facilities, high performance computing resources and domain knowledge to create training datasets at this scale and quality.

**Terabyte-scale X-ray microtomography images of the renal vascular network.** We acquired SRµCT images of whole mouse kidneys with 1.6 µm voxel size, ensuring sufficient sampling of functional capillaries, the smallest of which are 4 µm in diameter.[17] High image quality in terms of signal-to-noise ratio was achieved by employing the ID19 micro tomography beamline of the European Synchrotron Radiation Facility (ESRF), which provides several orders of magnitude higher brightness than conventional laboratory source microtomography devices. This allows for propagation-based phase contrast imaging, which leverages sample refractive index-dependent edge-enhancement for improved contrast.[18] Figure 1 provides an overview of the data acquisition and processing pipeline. Therein, the raw SRµCT radiographs are marked as **D1**.

Vascular signal-to-noise ratio was further improved by the application of a custom-developed mixture of contrast agent in a vascular casting resin capable of entering the smallest capillaries and filling the entire vascular bed. There was sufficient contrast to extract an initial blood vessel segment using curvelet-based denoising and hard thresholding. Connectivity analysis ensured and confirmed that the blood vessel segment, including capillary bed, was fully connected. Microscopic gas bubbles caused by external perfusion of the kidney, which is part of the organ preparation procedure, were excluded from the image set by applying connected component analysis to the background. This allowed for the elimination of all gas bubbles not in contact with the vessel boundary (**D5** in Figure 1).

The contrast agent, 1,3-diiodobenzene, also diffused into the lipophilic white adipose tissue, resulting in high X-ray absorption in the perirenal fat, which is, therefore, included in the vascular segment (Figure 2a). To visualize the segment without fat, a machine learning approach based on invariant scattering convolution networks[19] was applied, removing most of the blob-shaped fat globules and the more dense fatty tissue surrounding the collecting duct (**D5b**). Using the resulting segment, vessel thickness was calculated according to the concept of largest inscribed sphere (Figure 2b, d).

**Post-processed dataset and reference segmentations.** Ring artefacts compromise segmentation, as the corresponding areas may be erroneously attributed to the blood vessel or tubule segments, depending on their intensity (Figure 3a). To avoid this, kidneys were scanned with overlapping height steps, and the overlapping regions of the neighboring height steps were employed to produce a post-processed 3D volume with greatly reduced ring artefacts (**D3** in Figure 1), on which blood vessel and tubular segmentations were performed. Connectivity analysis yielded the fully connected tree of the blood vessel segment (**D5**). Insufficient contrast and resolution within the inner medulla prevented similar connectivity analysis on the tubular segment (**D6**).

**High quality annotations of glomeruli in full 3D.** Glomeruli are the primary filtration units of the kidney. They feature characteristic ball-shaped vascular structures. As they present with the same gray values and vessel diameters as other blood vessels in the kidney, differ only in their morphology, possess diverse sizes and shapes and are present in large, discrete numbers in the kidney, they pose a difficult segmentation challenge with a large sample size.

When contouring features of interest in 3D, however, the required workload is multiplied by the number of slices in the third dimension. In our work, this corresponds to a factor of 256 for a region of interest and 7168 for an entire kidney. Due to this extremely high contouring workload, 3D data are generally not annotated in full 3D, but rather partially annotated by selecting and contouring only a few sparse 2D slices out of the whole dataset.[7] This approach decreases the accuracy of the segmentation in the third dimension, which may be acceptable for clinical images, where standardized body positions or anisotropic resolutions reduce the necessity of rotational invariance in both algorithms and segmentation. In more generalized segmentation problems encountered in biomedical imaging on the other hand, this approach prevents these invariances from being leveraged for data augmentation techniques and may reduce performance of corresponding algorithms.[20] For this reason, the annotator manually contoured all slices in three full 3D regions of interest of 512 × 256 × 256 voxels in size (**D4** in Figure 1, Figure 3b).

**Identification of all individual glomeruli.** Combining these manual training data with the scattering transform approach, we were able to identify 10 031, 11 238 and 12 460 glomeruli, respectively, in the three kidney datasets. Each glomerulus was 3D visualized in a gallery (Figure 4b), viewed by a single domain expert and classified by its shape as false positive, true positive with shape distortion, or true positive without shape distortion. Only 15, 4 and 4 glomeruli, respectively, were identified as false positives by the rater, and were typically the result of poor contrast-to-noise in the specific region of the underlying raw images. Such localized areas of poor contrast appear to be caused by limited diffusion of the radiopaque 1,3-diiodobenzene from the vascular cast into the surrounding tissue, increasing background gray values (Figure 3c). False negatives were estimated to be approximately 2603, 2306 and 1420 using unbiased stereological counting on selected virtual sections, corresponding to miss rates of 20 %, 17 % and 10 %, respectively. It should be noted that both the manual segmentation of training data and validation were performed by the same, single annotator, meaning that annotator-specific bias cannot be excluded. On the other hand, the lack of inter-rater variability removes a source of label noise, which may be beneficial in some applications, such as allowing for a higher performance ceiling in benchmarking.[21]

The binary masks containing these glomeruli represent the supervised and validated training dataset (**D8** in Figure 1) and are available in the repository, along with the visualizations and expert classifications. Morphometric analysis was employed to separate individual glomeruli within clusters in which they appeared fused due to shared vessels (Figure 4c).

## Methods

Materials list, in-depth surgery guide and more detailed description of the data processing are provided in the Supplementary Information.

**Mouse husbandry.** C57BL/6J mice were purchased from Janvier Labs (Le Genest-Saint-Isle, France) and kept in individually ventilated cages with *ad libitum* access to water and standard diet (Kliba Nafag 3436, Kaiseraugst, Switzerland) in 12 h light/dark cycles. Dataset 1 derives from the left kidney of a male mouse, 15 weeks of age with a body weight of 28.0 g. Dataset 2 is the right kidney of the same mouse. Dataset 3 derives from the right kidney of a female mouse, 15 weeks of age with a body weight of 22.5 g. All animal experiments were approved by the cantonal veterinary office of Zurich, Switzerland, in accordance with the Swiss federal animal welfare regulations (license numbers ZH177/13 and ZH233/15).

**Perfusion surgery.** Mice were anaesthetized with ketamine/xylazine. A blunted 21G butterfly needle was inserted retrogradely into the abdominal aorta and fixed with a ligation (Figures 6, 7).[22] The abdominal aorta and superior mesenteric artery above the renal arteries were ligated, the vena cava opened as an outlet and the kidneys were flushed with 10 ml, 37 °C phosphate-buffered saline (PBS) to remove the blood, then fixed with 50 ml 37 °C 4 % paraformaldehyde in PBS (PFA) solution at 150 mmHg hydrostatic pressure.

**Vascular casting.** 2.4 g of 1,3-diiodobenzene (Sigma-Aldrich, Schnelldorf, Germany) were dissolved in 7.5 g of 2-butanone (Sigma-Aldrich) and mixed with 7.5 g PU4ii resin (vasQtec, Zurich, Switzerland) and 1.3 g PU4ii hardener. The mixture was filtered through a paper filter and degassed extensively in a vacuum chamber to minimize bubble formation during polymerization, and perfused at a constant pressure of no more than 200 mmHg until the resin mixture solidified. Kidneys were excised and stored in 15 ml 4 % PFA. For scanning, they were embedded in 2 % agar in PBS in 0.5 ml polypropylene centrifugation tubes. Kidneys were quality-checked with a nanotom® m (phoenix|x-ray, GE Sensing & Inspection Technologies GmbH, Wunstorf, Germany). Samples showing insufficient perfusion or bleeding of resin into the renal capsule or sinuses were excluded.

**ESRF ID19 micro-CT measurements.** Ten kidneys were scanned at the ID19 tomography beamline of the European Synchrotron Radiation Facility (ESRF, Grenoble, France) using pink beam with a mean photon energy of 19 keV. Radiographs were recorded at a sample-detector distance of 28 cm with a 100 μm Ce:LuAG scintillator, 4× magnification lens and a pco.edge 5.5 camera with a 2560 × 2160 pixel array and 6.5 μm pixel size, resulting in an effective pixel size of 1.625 μm. Radiographs were acquired with a half-acquisition scheme[23] in order to extend the field of view to 8 mm. Six height steps were recorded for each kidney, with half of the vertical field of view overlapping between each height step, resulting in fully redundant acquisition of the inner height steps.

5125 radiographs were recorded for each height step with 0.1 s exposure time, resulting in a scan time of 1 h for a whole kidney. 100 flat-field images were taken before and after each height step for flat-field correction. Images were reconstructed using the beamline's in-house PyHST2 software, using a Paganin-filter with a low $\delta/\beta$ ratio of 50 to limit loss in resolution and appearance of gradients close to large vessels.[18,24,25] Registration for stitching two half-acquisition radiographs to the full field of view was performed manually with 1 pixel accuracy. Data size for the reconstructed datasets was 1158 GB per kidney.

**Image stitching and inpainting.** Outliers in intensity in the recorded flat fields were segmented by noise reduction with 2D continuous curvelets, followed by thresholding to calculate radius and coordinates of the ring artefacts. The redundant acquisition of the central four height steps allowed us to replace corrupted data with a weighted average during stitching. The signals of the individual slices were zeroed in the presence of the rings, summed up and normalized by counting the number of uncorrupted signals. In the outer slices, where no redundant data was available, and in locations where rings coincided in both height steps, we employed a discrete cosine transform-based inpainting technique with a simple iterative approach, where we picked smoothing kernels progressively smaller in size and reconstructed the signal in the target areas by smoothing the signal everywhere at each iteration. The smoothed signal in the target areas was then combined with the original signal elsewhere to form a new image. In the next iteration, in turn, the new image was then smoothed to rewrite the signal at the target regions. The final inpainted signal exhibits multiple scales since different kernel widths are considered at different iterations.

The alignment for stitching the six stacks was determined by carrying out manual 3D registration and double checking against pairwise stack-stack phase-correlation analysis.[26] The stitching process reduced the dataset dimensions per kidney to 4608 × 4608 × 7168 pixels, totaling 567 GB.

**Semi-automatic segmentation of the vascular and tubular trees.** We performed image enhancement based on 3D discretized continuous curvelets,[27] in a similar fashion as Starck et al.,[28] but with second generation curvelets (i.e., no Radon transform) in 3D. The enhancement was carried out globally by leveraging the Fast Fourier Transform with MPI-FFTW,[29] considering about 100 curvelets. The "wedges" (curvelets in the spectrum) have a conical shape and cover the unit sphere in an approximately uniform fashion. For a given curvelet, a per-pixel coefficient is obtained by computing an inverse Fourier transform of its wedge and the image spectrum. We then truncated these coefficients in the image domain against a hard threshold, and forward-transformed the curvelet again into the Fourier space, modulated the curvelets with the truncated coefficients and superposed them. As a result, the pixel intensities were compressed to a substantially smaller range of values, thus helping to avoid over- and under-segmentation of large and small vessels, respectively. A threshold-based segmentation followed the image enhancement. The enhancement parameters and threshold were manually chosen by examining six randomly chosen regions of interest. Spurious islands were removed by 26-connected component analysis, and cavities were removed by 6-connected component analysis.

The bulk of the processing workload, required to transform data into an actionable training set, was carried out at the Zeus cluster of the Pawsey supercomputing centre. Zeus consisted of hundreds of computing nodes featuring Intel Xeon Phi (Knights Landing) many-core CPUs, together with 96 GB of ``special'' high-bandwidth memory (HBM/MCDRAM), as well as 128 GB of conventional DDR4 RAM. The final training and assessments were carried out at the Euler VI cluster of ETH Zurich, with two-socket nodes featuring AMD EPYC 7742 (Rome) CPUs and 512 GB of DDR4 RAM.

**Identification of glomeruli via scattering transform.** A machine learning-based approach relying on invariant scattering convolution networks was employed to segment the glomeruli and remove perirenal fat from the blood vessel segment.[19] For the glomerular training data, three selected regions of interest of 512 × 256 × 256 voxels in size were selected from the cortical region of one kidney (dataset 2) and segmented by a single annotator by fully manual contouring in all slices. For the fat, manual work was reduced by providing an initial semiautomatic segmentation, which the manual annotation then corrected. The training data were supplemented by additional regions of interest that contained no glomeruli or fat at all, and thus did not require manual annotation. The manual annotations were then used to train a hybrid algorithm that relied on a 3D scattering transform convolutional network topped with a dense neural network. The scattering transform relied upon ad-hoc designed 3D kernels (Morlet's wavelet with different sizes and orientations) that uniformly covered all directions at different scales. In the scattering convolutional network, filter nonlinearities were obtained by taking the magnitude of the filter responses and convolving them again with the kernels in a cascading fashion. These nonlinearities are designed to be robust against small Lipschitz-continuous deformations of the image.[19]

In contrast to our curvelet-based image enhancement approach, we decomposed the image into cubic tiles, then applied a windowed (thus local) Fourier transform on the tiles by considering regions about twice their size around them. While it would have been possible to use a convolutional network based upon a global scattering transform, this would have produced a very large number of features that would have had to be consumed at once, leading to an intermediate footprint in the petabyte-scale, exceeding the available memory of the cluster.

The scattering transform convolutional network produced a stack of a few hundred scalar feature maps per pixel. If considered as a "fiber bundle",[30,31] the feature map stack is equivariant under the symmetry group of rotations (i.e., the stack is a regular representation of the 3D rotation group SO(3)). This property can be exploited by further processing the feature maps with a dense neural network with increased parameter sharing across the hidden layers, making the output layer-invariant to rotations.

## Data Records

The dataset is available at the Image Data Resource (IDR) repository at https://doi.org/10.17867/10000188 under accession number idr0147.[32] As per the repository's guidelines, all data is available in the OME-TIFF format, which features the ability to load downsampled versions of the image data, as well as viewing them on the repository's online image viewer. Raw X-ray radiographs (**D1**) are further provided as an attachment in the original ESRF data format as well, which are raw binary image files with a 1024 bytes header describing the necessary metadata to open the images. All radiographs provided in this format feature image dimensions of 2560 × 2160 pixels, 16-bit unsigned integer bit depth and little-endian byte order.

**File structure**
The HR-Kidney datasets deposited in IDR are collected under a main folder named "idr0147-kuo-kidney3d". The datasets of the three kidneys are collected in three subfolders "Kidney_1", "Kidney_2" and "Kidney_3". The filenames of all data provided in the folder "Kidney_2" are provided in Table 1. Datasets of the other kidneys follow the same naming scheme, differing only in the kidney number.
Raw X-ray radiographs (**D1**) are provided both in the OME-TIFF-format, and in the original ESRF data format (.edf) as attachments on IDR. Manual annotation data (**D4**, **D9**) and defatted blood vessel segment (**D5b**) are only available for "Kidney_2". Annotation data are provided along with the excerpts (**D4**) from the reconstructed, Paganin-filtered 3D volumes (**D2**), denoted with a filename ending with "_pag". Excerpts of the ROIs of the equivalent 3D volumes reconstructed without Paganin-filtering ("absorption images") are provided with a filename ending with "_abs", as they formed the basis of the manual annotation. They were not employed for training and are only provided for documentation. As the two versions are reconstructions of the same dataset, the manual annotations are equally valid for both. Coordinates and dimensions for extracting the ROIs are provided as metadata on IDR.

# Technical Validation

**Validation of glomeruli by a domain expert.** To reduce the workload to the level required to make validation of each glomerulus feasible, a volumetric visualization of the overlap between the blood vessel mask and the glomerular mask was generated for rapid evaluation of easy to recognize glomeruli or artefacts. In a first round, glomeruli were classified by their shape into three categories: 1. certain true positive with shape distortion, 2. certain true positive without shape distortion and 3. uncertain. Candidates of the third category, which constituted about 3 to 5 % of all machine-identified glomeruli, were then reviewed in a second round as an overlay over the original raw data on a slice-by-slice basis. These candidates were then assigned as false positives or as certain glomeruli of categories 1 or 2.

The numbers of false negative rates were assessed using stereological counting, to ensure proper unbiased sampling.[33] Four pairs of virtual sections were selected at equidistant intervals in one half of each kidney. The distance between the pairs was 31 slices or 50 µm. To extrapolate the number of counted glomeruli to the whole kidney, both the considered volume and the volume of the entire kidney were calculated based on a mask derived from morphological closing on the vascular segmentation (**D5**).

The vascular and tubular structures do not form individual, countable units. Therefore, the accuracy of their semi-automatic segmentations (**D5**, **D6**) cannot be assessed on a similar per-object-basis. An assessment would have to be performed on each individual voxel instead. This would require voxel-accurate ground truth data with higher precision than can be created, with reasonable workload, using manual annotation of the highly intricate and convoluted vascular and tubular trees. Accordingly, these segmentations have not been validated to be used as ground truth for machine-learning or benchmarking datasets, but are rather supplied as examples of segmentations for other purposes, such as testing image processing, segmentation or vascular analysis methods at the terabyte scale.

# Code Availability

The HR-Kidney dataset is freely available for download at the Image Data Resource under accession number idr0147:

https://doi.org/10.17867/10000188

# Acknowledgements
We gratefully acknowledge access to and support at the Swiss National Supercomputing Centre (Maria Grazia Giuffreda) and the Pawsey Supercomputing Centre (Ugo Varetto, David Schibeci). We acknowledge the European Synchrotron Radiation Facility (ESRF) for provision of synchrotron radiation facilities under proposal numbers MD-1014 and MD-1017 and we would like to thank Alexander Rack and Margie Olbinado for assistance in using beamline ID19. We wish to thank Axel Lang, Erich Meyer, Anastasios Marmaras and Virginia Meskenaite for help with vascular casting. Jan Czogalla for help with establishing the perfusion surgery. Gilles Fourestey for support in high performance computing, Lucid Concepts AG for collaboration in software development. Michael Bergdorf for help with scattering transform schemes. We are grateful for the financial support provided by the Swiss National Science Foundation (grant 205321_153523, to V.K., S.H. and B.M.) and NCCR Kidney.CH (grant 183774, to V.K. and R.H.W.).

## Author contributions
W.K. developed the vascular resin mixture and surgery protocol, and performed SRµCT imaging, reconstructions, manual annotation, validation, data curation and analysis. D.R. developed the pipeline and performed the processing for artefact-correction, segmentations, machine learning, path calculations, visualizations and data analysis. W.K. and D.R. drafted the manuscript. G.S. and S.H. performed SRµCT imaging and designed scanning parameters. R.H.W., S.H., B.M. and V.K. conceived and designed the research, and edited and revised the manuscript.

## Competing interests
The authors declare no competing interests.

# Figures

**Figure 1**

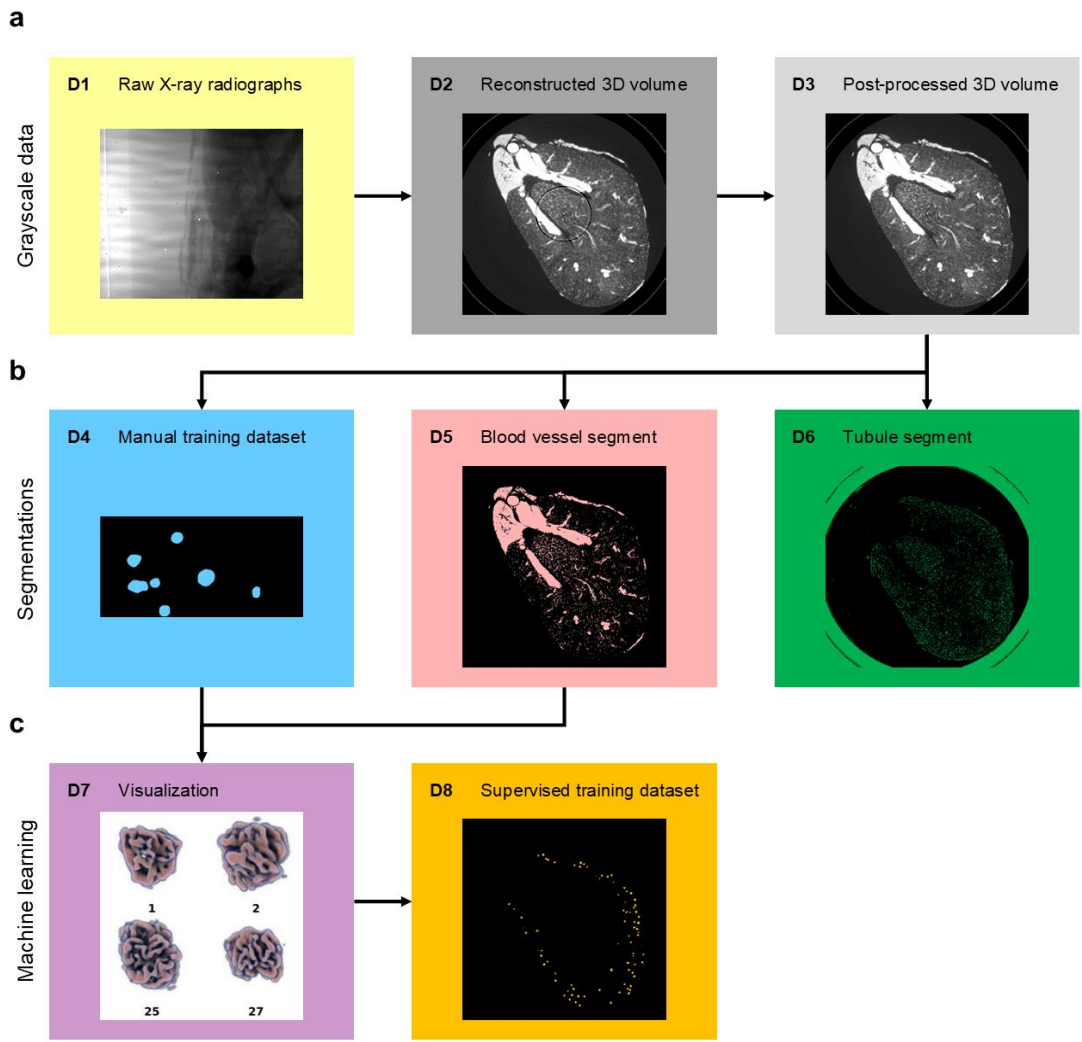

*Figure 1: Overview of the data made available in the repository. (**a**): Grayscale data: X-ray radiographs (**D1**) were reconstructed into 3D volumes (**D2**) and post-processed for ring artefact removal (**D3**). (**b**): Classical segmentations: Glomeruli were manually contoured in small regions of interests (**D4**). Blood vessels (**D5**) and tubules (**D6**) were segmented by noise removal, thresholding and connectivity analysis. (**c**): Machine learning segmentations: Glomeruli identified via machine learning were combined with the blood vessel segment for visualization (**D7**), which was viewed by a domain expert to validate the final glomerulus mask (**D8**). A list of dataset contents and file formats is provided in Table 1.*

**Figure 2**

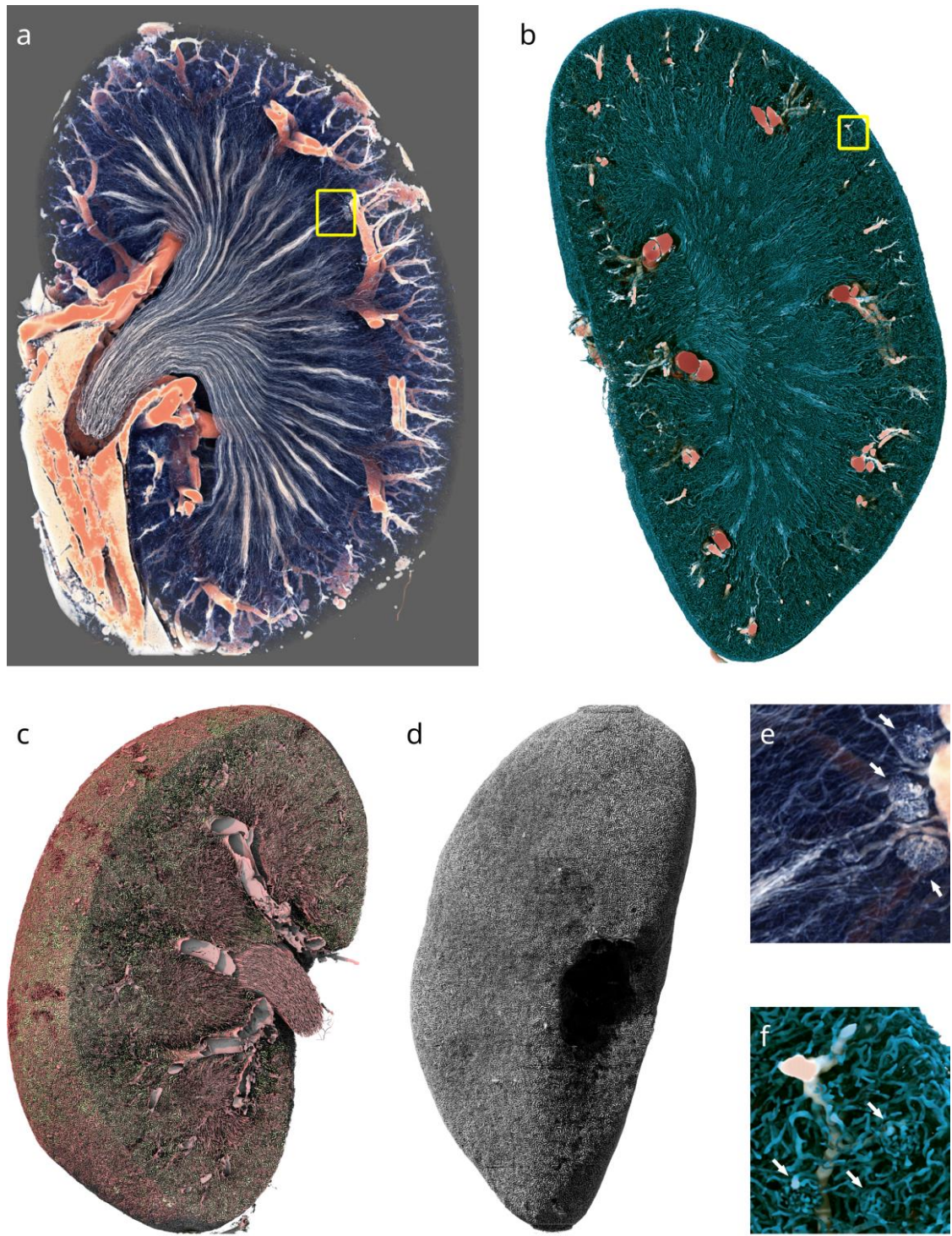

*Figure 2: Computer graphics renderings of the vascular and tubular structure of a mouse kidney. (**a**): Volume rendering of a Paganin-filtered kidney dataset (**D2**), clipped for visibility. Red colors represent higher intensity. Only the higher intensity values, which are due to the contrast agent in larger blood vessels and perirenal fat, are shown. (**b**): Opaque rendering of the thickness transform of the defatted blood vessel binary mask (**D2b**). Colors correspond to largest inscribed sphere radius. (**c**): Surface rendering of the post-processed segmented vascular (red) and tubular (green) lumina in the cortex and inner medulla. D: Surface rendering of the segmented tubular lumina only (**D6**). E: Magnified view of the region highlighted by the yellow square in Figure 2A. Three juxtamedullary glomeruli are visible in*

*the right part of the image (white arrows). F: Magnified view of the region highlighted by the yellow square in Figure 2B. Three cortical glomeruli are visible in the bottom half of the image (white arrows). Height of the kidney: 10 mm.*

**Figure 3**

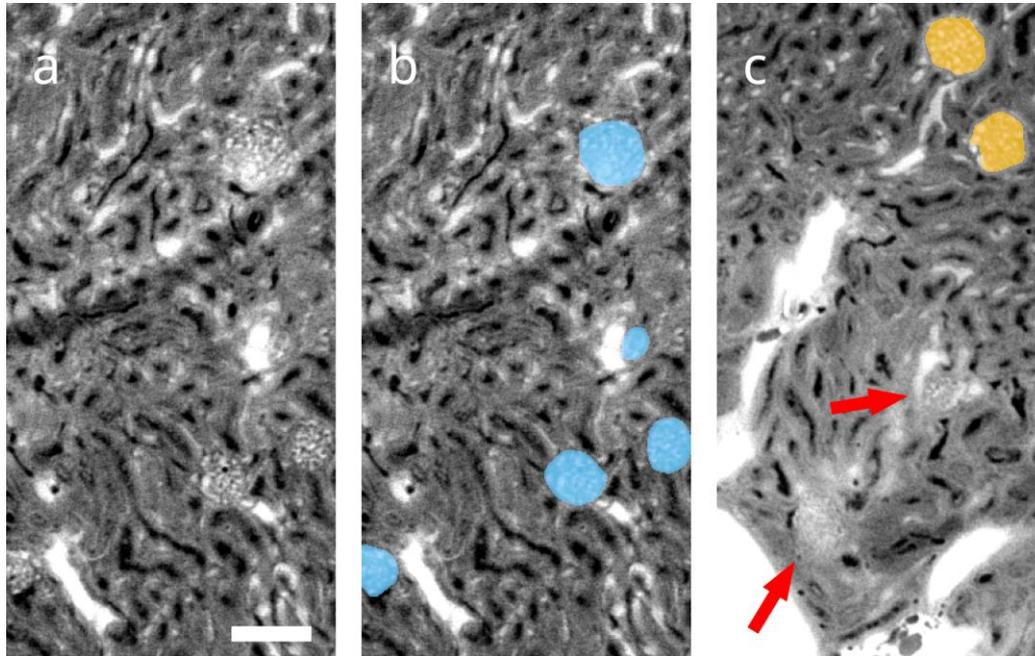

Figure 3. *Local views of the reconstructed 3D volume prior to artefact correction (**D2**). (**a**): A single virtual section of one of the regions of interest selected for manual contouring. Glomerular blood vessels feature the same gray values and size scales as other blood vessels in the kidney and differ only in their morphology. A prominent ring artefact can be observed touching the top glomerulus. Scale bar: 100 µm (**b**): Manual annotation (**D4**) of glomeruli overlayed in blue over raw data. (**c**): Different region of interest containing glomeruli identified by machine learning (**D8**) overlayed in orange over raw data, as well as false negatives denoted by red arrows. The majority of missed glomeruli are in similar regions of poor contrast-to-noise, which are characterized by elevated tissue background intensity. These are likely caused by limited diffusion of 1,3-diiodobenzene from the cast into surrounding tissue. The few false positives identified are also located mainly in such regions.*

**Figure 4**

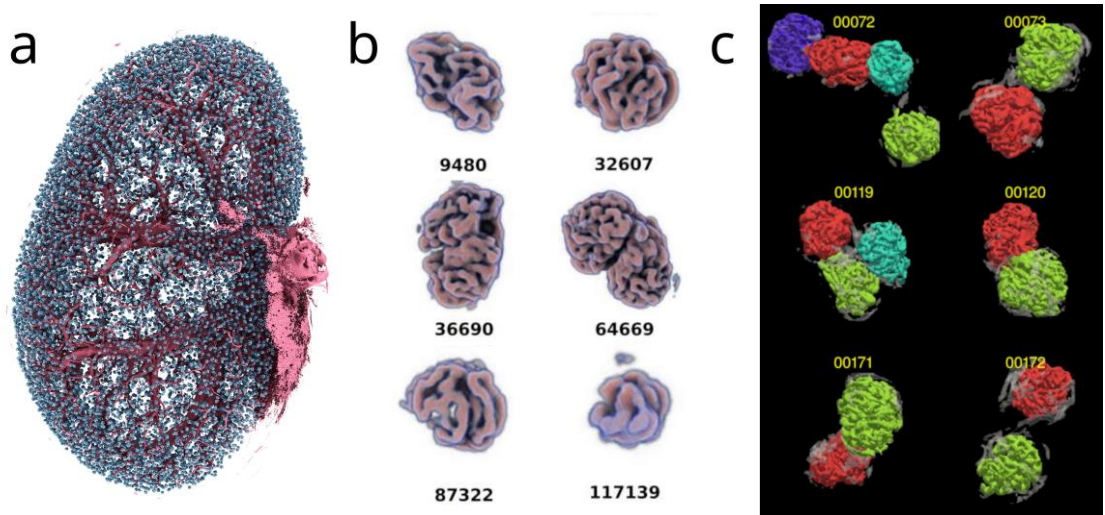

*Figure 4: Computer generated images of glomeruli. (**a**): All identified glomeruli (**D8**) of one kidney are shown in cyan in their original spatial location. Large pre-glomerular vessels are rendered in magenta, for orientation. (**b**): Excerpt from the gallery of volume-rendered glomeruli, exhibiting a selection of different glomerulus sizes and shapes. Each glomerulus or cluster of glomeruli was assigned an identification number and viewed by an expert. (**c**): Results of the morphometric analysis to separate clustered glomeruli.*

# Tables

**Table 1**
List of filenames, data contents, data types, bit depths and data formats contained within the HR-Kidney dataset available at the Image Data Resource (IDR) repository. Filenames are indicated for "Kidney_2", which is the only dataset for which manual annotation data (D4, D9) and defatted blood vessel segment (D5b) are available.

|  | **Filenames (Kidney_2)** | **Data provided** | **Precision** | **Data format** |
|---|---|---|---|---|
| D1 | kidney2_D1_projections_height1.ome.tiff<br>kidney2_D1_projections_height2.ome.tiff<br>kidney2_D1_projections_height3.ome.tiff<br>kidney2_D1_projections_height4.ome.tiff<br>kidney2_D1_projections_height5.ome.tiff<br>kidney2_D1_projections_height6.ome.tiff | X-ray radiographs | Grayscale, 16-bit unsigned integer | OME-TIFF (.tif)<br><br>In attachment:<br>ESRF data format (.edf, raw binary with 1024 byte header describing image dimensions) |
| D2 | kidney2_D2_reco_height1.ome.tiff<br>kidney2_D2_reco_height2.ome.tiff<br>kidney2_D2_reco_height3.ome.tiff<br>kidney2_D2_reco_height4.ome.tiff<br>kidney2_D2_reco_height5.ome.tiff<br>kidney2_D2_reco_height6.ome.tiff | Paganin filtered, reconstructed 3D volume | Grayscale, 32-bit floating point | OME-TIFF (.tif) |
| D3 | kidney2_D3_inpainted.ome.tiff | Artifact-corrected, inpainted 3D volume | Grayscale, 32-bit floating point | OME-TIFF (.tif) |
| D4 | kidney2_D4_h3_glomeruli_roi1_annotation.ome.tiff<br>kidney2_D4_h3_glomeruli_roi1_raw_abs.ome.tiff<br>kidney2_D4_h3_glomeruli_roi1_raw_pag.ome.tiff<br>kidney2_D4_h4_glomeruli_roi2_annotation.ome.tiff<br>kidney2_D4_h4_glomeruli_roi2_raw_abs.ome.tiff<br>kidney2_D4_h4_glomeruli_roi2_raw_pag.ome.tiff<br>kidney2_D4_h5_glomeruli_roi3_annotation.ome.tiff<br>kidney2_D4_h5_glomeruli_roi3_raw_abs.ome.tiff<br>kidney2_D4_h5_glomeruli_roi3_raw_pag.ome.tiff | Glomeruli annotation, 3 regions of interests (512 × 256 × 256) | Binary, 8-bit integer | OME-TIFF (.tif) |
| D5 | kidney2_D5_blood_vessels.ome.tiff | Blood vessel segment | Binary, 8-bit integer | OME-TIFF (.tif) |
| D5b | kidney2_D5b_blood_vessels_defatted.ome.tiff | Defatted blood vessel segment | Binary, 8-bit integer | OME-TIFF (.tif) |
| D6 | kidney2_D6_tubules.ome.tiff | Tubule segment | Binary, 8-bit integer | OME-TIFF (.tiff) |
| D7 | kidney2_D7_glomgallery_00.ome.tiff<br>kidney2_D7_glomgallery_01.ome.tiff<br>kidney2_D7_glomgallery_02.ome.tiff<br>kidney2_D7_glomgallery_03.ome.tiff<br>kidney2_D7_glomgallery_04.ome.tiff<br>kidney2_D7_glomgallery_05.ome.tiff<br>kidney2_D7_glomgallery_06.ome.tiff<br>kidney2_D7_glomgallery_07.ome.tiff<br>kidney2_D7_glomgallery_08.ome.tiff<br>kidney2_D7_glomgallery_09.ome.tiff<br>kidney2_D7_glomgallery_10.ome.tiff<br>kidney2_D7_glomgallery_11.ome.tiff | Gallery of visualized glomeruli | RGB, 8-bit integer per channel | OME-TIFF (.tiff) |
| D8 | kidney2_D8_glomeruli_segment.ome.tiff | Glomeruli segment | Binary, 8-bit integer | OME-TIFF (.tiff) |
| D9 | kidney2_D9_h2_fat_roi1_annotation.ome.tiff<br>kidney2_D9_h2_fat_roi1_raw_abs.ome.tiff<br>kidney2_D9_h2_fat_roi1_raw_pag.ome.tiff<br>kidney2_D9_h2_fat_roi2_annotation.ome.tiff<br>kidney2_D9_h2_fat_roi2_raw_abs.ome.tiff<br>kidney2_D9_h2_fat_roi2_raw_pag.ome.tiff<br>kidney2_D9_h4_fat_roi3_annotation.ome.tiff<br>kidney2_D9_h4_fat_roi3_raw_abs.ome.tiff<br>kidney2_D9_h4_fat_roi3_raw_pag.ome.tiff | Extrarenal fat annotation, 3 regions of interests (512 × 256 × 256) | Binary, 8-bit integer | OME-TIFF (.tiff) |

# Supplementary Information

## Terabyte-scale supervised 3D training and benchmarking dataset of the mouse kidney

### Authors

Willy Kuo[1,2,*], Diego Rossinelli[1,2,*], Georg Schulz[3], Roland H. Wenger[1,2], Simone Hieber[3], Bert Müller[3,†], Vartan Kurtcuoglu[1,2,†]

### Affiliations

1. Institute of Physiology, University of Zurich, Zurich, Switzerland

2. National Centre of Competence in Research, Kidney.CH, Zurich, Switzerland

3. Biomaterials Science Center, Department of Biomedical Engineering, University of Basel, Allschwil, Switzerland

* These authors contributed equally to this work

† These authors jointly supervised this work

Corresponding author: Vartan Kurtcuoglu (vartan.kurtcuoglu@uzh.ch)

## Table of Contents

# Materials and Suppliers List

**Perfusion reagents**

| Phosphate Buffered Saline (PBS) | Oxoid Phosphate Buffered Saline Tablets (Dulbecco A) <br><br> BR0014G, ThermoFisher Scientific, United States |
|---|---|
| Ketamine 100 mg/ml | Ketasol®-100 ad us. vet., injection solution <br><br> Dr. E. Graeub AG, Switzerland |
| Xylazine 20 mg/ml | Xylazin Streuli ad us. vet., injection solution <br><br> Streuli Pharma AG, Switzerland |
| Paraformaldehyde | Paraformaldehyde prilled, 95% <br><br> 441244, Sigma Aldrich, Germany |
| Mineral oil | Mineral oil, light oil (neat), BioReagent <br><br> M8410, Sigma Aldrich, Germany |

**Surgery Tools**

| Fine scissors | Vannas Spring Scissors - 2.5mm Blades <br><br> 15000-08, Fine Science Tools, Germany |
|---|---|
| Arterial clamp | Micro Serrefine - $10 \times 2$ mm <br><br> 18055-01, Fine Science Tools, Germany |
| Arterial clamp applying forceps | Micro-Serrefine Clip Applying Forceps <br><br> 18057-14, Fine Science Tools, Germany |
| Vessel dilating forceps | S&T Vessel Dilating Forceps - 11cm <br><br> 00125-11, Fine Science Tools, Germany |
| Angled forceps | S&T 0.3mm $\times$ 0.25mm Forceps <br><br> 00649-11, Fine Science Tools, Germany |
| Straight forceps | Rubis Switzerland Tweezers 5-SA <br><br> 232-1221, VWR, United States |
| Silk suture for ligations | Non-Sterile Silk Suture Thread 5/0 <br><br> 18020-50, Fine Science Tools, Germany |

**Perfusion Consumables**

| | |
|---|---|
| 1 ml syringe | Injekt F 1 ml |
| | 9166017V, B. Braun, Germany |
| 26 G needle | Sterican 26 G × ½ " |
| | 466 5457, B. Braun, Germany |
| 10 ml syringe Luer Lock | NORM-JECT 10 ml (12 ml) |
| | 4100-X00V0, Henke Sass Wolf, Germany |
| 50 ml syringe Luer Lock | Omnifix 50 ml (60 ml) |
| | 4617509F , B. Braun, Germany |
| 1.2 μm syringe filter | Chromafil Xtra PET-120/25 |
| | 729229, Macherey-Nagel, Germany |
| 3-way stopcock | Discofix C 3-way Stopcock |
| | 16494C, B. Braun, Germany |
| 21 G butterfly needle | Venofix Safety G21 |
| | 4056521-01, B. Braun, Germany |
| 0.5 ml centrifugation tubes | PCR Single tubes, PP, 0,5 ml |
| | 781310, Brand, Germany |
| 1.5 ml centrifugation tubes | Micro tube 1.5ml |
| | 72.690.001, Sarstedt, Germany |

**Vascular casting**

| | |
|---|---|
| PU4ii | PU4ii |
| | VasQtec, Switzerland |
| 1,3-Diiodobenzene | 1,3-Diiodobenzene |
| | 475262, Sigma Aldrich, Germany |
| 2-Butanone | 2-Butanone |
| | 360473, Sigma Aldrich, Germany |

# Detailed Methods

Optimized vascular casting technique

Standard transcardial perfusions as used for whole animal vascular casting are unable to fill the renal vascular tree in its entirety, as most of the resin will flow to the outlet via lower resistance paths, such as the liver. As a result, resin droplets enclosed within residual water can be observed in venous vessels (Supplemental Figure S1).

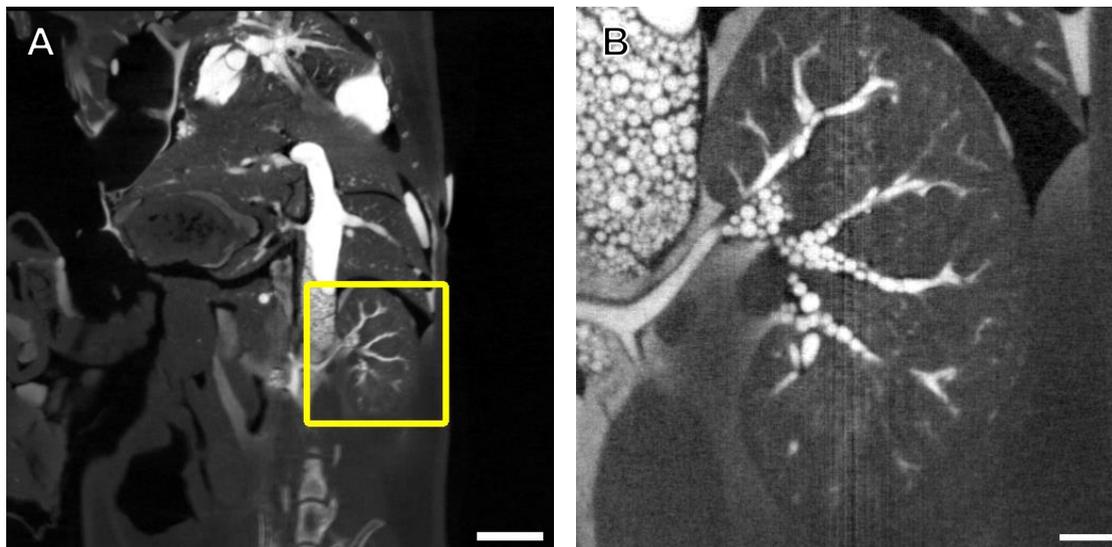

*Supplemental Figure S1: Results of an unoptimized transcardial vascular casting. Resin droplets enclosed in residual water can be observed in the renal venous vessel tree and the vena cava. A: Overview showing abdomen and thorax. Scale bar: 5 mm. B: Magnified view of the region of interest marked in yellow. Scale bar: 1 mm.*

Optimized perfusion techniques are required, where all resin flow is diverted to the organ of interest and lower resistance pathways are closed off via ligations.[1] The herein presented method is derived from the isolated perfused kidney technique[2] and was adapted for vascular casting.

Preparation

21 G butterfly needles were blunted by filing down their needle points with a metal file. Needles smaller than 21 G should not be employed for vascular casting, as they may not support the necessary flow rates. The outer surfaces of the needle were deburred with gentle filing in order to remove sharp edges or bumps, which may damage the vessel during needle insertion. The inside surfaces were deburred with a 26 G needle tip, in order to prevent flow restrictions.

The butterfly needle was then connected to a 3-way stopcock for flow control, which in turn was connected to 2.5 m long silicon tubing via Luer lock connector. A 50 ml syringe was connected to the silicone tubing to serve as a reservoir. The contraption was flushed with water until all air bubbles were removed, then filled with phosphate-buffered saline (PBS). The reservoir was hung at 2 m height above the working space to provide 150 mmHg of hydrostatic pressure.

2-Butanone should be degassed before use. Per mouse, 2.4 g of 1,3-diiodobenzene were dissolved in 7.5 g of 2-butanone and mixed with 7.5 g PU4ii resin within 50 ml centrifugation tubes in preparation. This corresponds to an iodine concentration of 100 mg iodine / g in the final vascular cast. PBS and formaldehyde solutions (PFA) were kept in a 37 °C water bath.

Preparing ligations for diverting flow

Mice were anaesthetized with 120 mg/kg ketamine and 24 mg/kg xylazine, with additional doses of 25 mg/kg ketamine and 4 mg/kg xylazine given after 15 min if surgical tolerance was not achieved. The animal was fixed to a foam board, which would allow fixation of the butterfly needle with pins at a later stage. The abdomen was opened with a vertical cut, and the intestine was moved to the right side (from the observer's point of view). Throughout the entire surgery, the kidneys were wetted with 37° C PBS to prevent drying and premature coagulation. A part of the mesentery, visible as small transparent membrane, connects the intestine to the liver. This membrane was cut, so that the intestine could be moved entirely out of the way to the right without tearing the liver. A small needle cap was put under the back of the mouse to push up the region containing the superior mesenteric artery (SMA, Supplemental Figure S2). This vessel can be identified by the white tubing surrounding it, which is part of the mesentery, and should not be confused with venous vessels embedded in the surrounding fatty tissue. Fatty tissue does not form a distinct tubing and is more yellowish-white in color. The superior mesenteric artery can then be followed to find the proper ligation position at the abdominal aorta, which is anterior to the T-section where the superior mesenteric artery branches off of the abdominal aorta (AA, Supplemental Figure S2).

Fatty tissue around the ligation point was removed so that a 5/0 silk suture greased with mineral oil could be passed below the abdominal aorta. A constrictor knot ligation was prepared, but not closed (L1, Supplemental Figure S2). A second constrictor knot ligation was prepared at the mesenteric artery (L2, Supplemental Figure S2). The standard surgeon's throw may not be used for these ligations, as it has been found to open at the pressures used in the vascular casting procedure. In an *in vitro* model, it has been found to leak at pressures as low as 33 mmHg, whereas the constrictor knot holds sufficiently tight at perfusion pressures of 363 mmHg under the same conditions.[3]

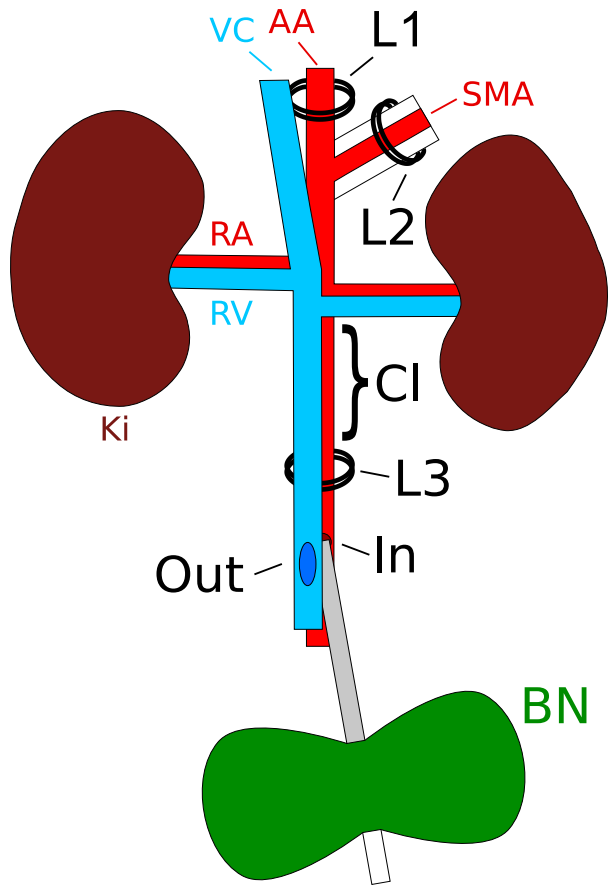

*Supplemental Figure S2: Schematic overview of the surgery steps necessary to perfuse kidneys (Ki). Constrictor knot ligations (L1, L2) around the abdominal aorta (AA) and superior mesenteric artery (SMA) were prepared. A third ligation (L3) was prepared posterior to the renal arteries (RA) and renal veins (RV). The abdominal aorta was clamped (Cl), and an incision was made to insert a blunted butterfly needle (In, BN), which was then fixed in place with the prepared ligation (L3). The vessel clamp (Cl) was removed, the remaining ligations (L1, L2) were closed and an incision was made into the vena cava (VC) to serve as an outlet (Out).*

Preparing ligation for needle insertion
The ideal needle insertion point (In, Supplemental Figure 2) in the abdominal aorta is posterior to the renal arteries (RA, Supplemental Figure 2). The longer the vessel segment between the insertion point and the renal arteries, the more securely the needle can be held in place by the ligation (L3, Supplemental Figure 2). In addition, more space allows additional attempts of inserting the needle to be taken further upstream the vessel if necessary. In practice, the needle ligation point will have to be chosen as a compromise based on where the abdominal aorta can be separated from the vena cava (VC, Supplemental Figure S2) for the ligation to be tied around the vessel, which is typically easier to perform closer to the renal arteries.

The vena cava can be readily identified as the largest vessel within the abdomen. Typically, it will be covered by fatty tissue, which should be split with tweezers to obtain access. The abdominal aorta can typically be found underneath the vena cava, towards the right side. Both of these blood vessels are surrounded by a fascia tubing, which needs to be cut or split with tweezers. The abdominal aorta can be grabbed gently with tweezers and pulled to the side without damaging the vessel. This may reveal a gap between the abdominal aorta and vena cava, where it's possible to cut through the fascia tubing without injuring either vessel. Note that the vena cava may not be grabbed at any point, as it is very likely to burst. While leakage through the vena cava does not disturb the vascular casting process later on, as it will opened as an outlet anyways, the resulting bleeding will disturb the operator's vision during the surgery until bleeding subsides, and lead to premature death of the animal.

If there is no gap large enough for surgical scissors to cut to allow separation of the two vessels, one should grab a part of the fascia tubing with one set of tweezers without grabbing either vessel, then grab onto the same part with a second set and gently pull the fascia tubing apart. Care needs to be taken to avoid exerting pulling force on the vena cava, as this may lead to vessel injury. This means that several small splits should be employed, rather than one big one. The gap created by this process may then be large enough to be cut through with surgical scissors, or may already be large enough on its own to pass through a greased silk suture.

A ligation was prepared around the abdominal aorta after the above process, but not closed yet (L3, Supplemental Figure S2). The abdominal aorta was then clamped between the ligation and the renal arteries to stop blood flow (Cl, Supplemental Figure S2). The vena cava may be partially clamped as well, in case that there is no full separation of the two vessels at this location.

Needle insertion and initial flushing
A small incision was made into the abdominal aorta with as much distance of the ligation as possible, to preserve as much vessel length as possible (In, Supplemental Figure S2). The incision should cut through 50 % of the vessel. Smaller cuts make it difficult to insert the vessel dilating forceps into the hole, larger cuts result in a more flaccid vessel that does not stay in place well during insertion.

Vessel dilating forceps were inserted into the hole, the vessel spread and the blunted 21 G butterfly needle (BN, Supplemental Figure S2) inserted between the forceps arms until it reached under the prepared ligation, and pushed against the vessel clamp. The butterfly needle was pinned behind its wings to prevent the needle to slip out of the vessel. The constrictor knot ligation (L3, Supplemental Figure S2) was then closed tight and the vessel clamp removed. The ligation of the superior mesenteric artery and ligation of the upper abdominal aorta were then closed as well.

A small window was cut into the vena cava some distance away from the renal veins to prevent them from collapsing under low pressure (Out, Supplemental Figure S2). For beginners, it is recommended to cut posterior to the ligation holding the needle in place (L3, Supplemental Figure S2), to avoid any consequences of accidentally cutting into the abdominal aorta. The kidneys were then flushed with 10 ml 37 °C PBS to remove the blood, then with 50 ml 37 °C 4 % formaldehyde in PBS (PFA) solution at 150 mmHg hydrostatic pressure. Flow rates can be observed based on the speed of drainage of the graduated syringe serving as reservoir, and typically achieve 5 ml / min during this step. This allows for a 10 min time window to prepare the PU4ii mixture, which can be extended by adding additional PFA solution to be perfused if required. Ideally, kidneys assume a pale coloration without visible blood patches after this step. (Supplemental Figure S3)

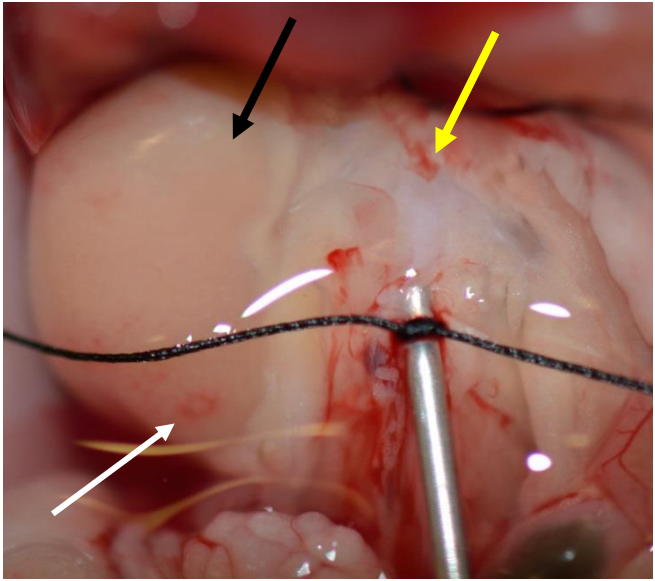

*Supplemental Figure S3: Right mouse kidney after perfusion with formaldehyde solution. Pale coloration of kidney and vena cava indicating successful flushing of blood (black and yellow arrows). One small blood patch can be observed on the bottom of the kidney (white arrow), which may indicate residual blood within the kidney, but may also be blood stuck to the outside of the organ as can be observed on top of the vena cava (yellow arrow).*

Vascular casting

1.3 g PU4ii hardener was added to the prepared mixture of 1,3-diiodobenzene, 2-butanone and PU4ii resin, and the mixture was filtered through a paper filter. The solution was degassed extensively in a vacuum chamber to minimize bubble formation during polymerization. It was then transferred to a 10 ml syringe with Luer lock. The 3-way stopcock was closed to stop flow, then the silicon tubing was disconnected and replaced with the resin-filled syringe.

The resin was perfused at a pressure of around 200 mmHg by actuating the syringe with a constant weight. The resin mixture was perfused until no more water bubbles could be observed exiting the outlet, which typically occurred after 3 ml of perfused volume. Typical flow rates are in the range of 0.2 ml / min during this step. If resin is observed exiting the abdominal aorta via the needle insertion point (In, Supplemental Figure S2), the ligation knot should be held tight with tweezers throughout the procedure (L3, Supplemental Figure S2). Pressure was kept throughout the entire procedure until the resin mixture solidified.

Supraphysiological perfusion pressures up to 200 mmHg cause distension of the blood vessel diameters, as well as mechanical compression of adjacent tubular luminae. Perfusion pressures above 200 mmHg may lead to bursting of the vessels and bleeding of vascular casting resin into the renal capsule, which results in distortion of the outer shape[4]. As such, perfusion pressure has to be chosen as a compromise between achieving reliable filling of the vasculature and staying close to physiological pressure ranges. It is, therefore, crucial to employ pressure-controlled pumps rather than standard syringe pumps with constant flow rates.

After polymerization, kidneys were excised and stored in 15 ml 4 % PFA. Kidneys were trimmed of fat as much as was possible without disturbing the outer surface of the kidney, as the radiopaque 1,3-diiodobenzene within the PU4ii mixture could diffuse into white adipose tissue. This issue can, in principle, be resolved by employing µAngiofil, another microparticle-free vascular casting reagent, which became available commercially after we concluded our experiments and which does not show this behavior.[4] We have, however, found that the recommended degassing procedure was insufficient to prevent the formation of large gas bubbles in the resulting cast.[5] Since viscosity and contrast are otherwise comparable to our mixture and µAngiofil features considerably higher materials cost (270 $ per mouse compared to 20 $), we did not repeat our experiments with the new vascular casting resin.

For scanning, kidneys were embedded in 2 % agar in PBS in 0.5 ml polypropylene centrifugation tubes with an inner diameter of 6 mm and an outer diameter of about 7.9 mm. The agar solution was degassed briefly in vacuum after heating to 100 °C, and left to cool down to approximately 40 °C before embedding the kidney. Care needs to be taken to avoid the introduction of any gas bubbles into the agar, as these will expand under the high radiation doses employed in this work's synchrotron radiation-based hard X-ray phase-contrast microtomography (SRµCT) imaging, resulting in movement artifacts.

ESRF ID19 SRµCT measurements
Kidneys were scanned at the ID19 tomography beamline of the European Synchrotron Radiation Facility (ESRF, Grenoble, France) using pink beam with a mean photon energy of 19 keV. Radiographs were recorded at a sample-detector distance of 28 cm with a 100 µm Ce:LuAG scintillator, 4× magnification lens and a pco.edge 5.5 camera with a 2560 × 2160 pixel array and 6.5 µm wide pixels, resulting in an effective pixel size of 1.625 µm and a field of view of 4.16 mm in width and 3.51 mm in height.

To extend the field of view, radiographs were acquired with an asymmetric rotation axis scheme: by offsetting the rotation axis, a single radiograph of half of the width of sample could be stitched with its 180° rotated equivalent of the other half. For registration, an approximately 200 pixel broad overlap was kept, resulting in extended field of view of 8 mm.

Six height steps were recorded for each kidney, with half of the vertical field of view overlapping between each height step. This resulted in fully redundant acquisition of the kidneys with the exception of the halves of the top and bottom height steps, which was later employed for ring artifact removal.

5125 radiographs were recorded for each height step with 0.1 s exposure time, resulting in a scan time of 10 min per height step or one hour for a whole kidney. This allowed us to remain below the time threshold at which gas bubbles would start to form within the agar, which was around the 15 min mark. 100 flat-field images were taken before and after each height step for flat-field correction. Data size of the radiographs was 55 GB per height step, or 330 GB per kidney (**D1**).

Images were reconstructed using the beamline's in-house PyHST2 software[6], using a Paganin-filter[7] with a low δ/β ratio of 50 to limit loss in resolution and appearance of gradients close to large vessels.[8] Registration for stitching two radiographs for the full field of view was performed manually with 1 pixel accuracy. All other parameters were kept on default settings. Data size of the reconstructed datasets was 193 GB per height step, or 1158 GB per kidney (**D2**).

Ring artifact removal
Physical phenomena such as dust on the scintillator and highly absorbing particles in the beam path can result in a deviation from the expected response of the detector, producing ring artifacts at those locations. Outliers in intensity in the recorded flat fields were segmented to calculate radius and coordinates of the ring artifacts by noise reduction with 2D continuous curvelets, followed by thresholding. The redundant acquisition of the central four height steps allowed us to replace corrupted data with a weighted average during stitching. The grey values of the individual slices were zeroed in the presence of the rings, summed up and divided by the number of uncorrupted signals. In the outer slices, where no redundant data was available, and in locations where rings coincided in both height steps, we employed a discrete cosine transform-based (DCT) inpainting technique with a simple iterative approach, where we picked Gaussian smoothing kernels progressively smaller in size and reconstructed the signal in the target areas by smoothing the signal everywhere at each iteration. The smooth signal in the target areas is then combined with the original signal elsewhere, to form a new image. In the next iteration, in turn, the new image is then smoothed to rewrite the signal at the target regions. The final inpainted signal exhibits multiple scales, since different kernel widths are considered at different iterations, starting with closing the gaps with the largest frequencies and progressive refinements with higher frequencies.

The alignment for stitching the six stacks was determined by carrying out manual 3D registration and double checking against pairwise stack-stack phase-correlation analysis.[9] The stitching process reduced the dataset dimension per kidney to 4608 × 4608 × 7168 pixels, totaling 567 GB (**D3**).

Continuous curvelets denoising
We performed image enhancement based on 3D discretized continuous curvelets[10], in a similar fashion as Starck et al.[11] but with second generation curvelets (i.e. no Radon transform) in 3D. The enhancement was carried out globally by leveraging the Fast Fourier Transform with MPI-FFTW[12], considering about 100 curvelets. The "wedges" (curvelets in the spectrum) have a conical shape and cover the unit sphere in an approximately uniform fashion. For a given curvelet, a per-pixel coefficient is obtained by computing an inverse Fourier transform of its wedge and the image spectrum. We then truncated these coefficients in the image domain against a hard threshold, and forward-transformed it again into the Fourier space, modulated the curvelets with the truncated coefficients and superposed them. The end effect of our approach has shown to squeeze the pixel intensities into a substantially more limited range of values, thus helping to avoid over- and under-segmentation of large and small vessels, respectively. A threshold-based segmentation followed the image enhancement. The enhancement parameters and threshold were manually chosen by examining six randomly chosen regions of interest. Spurious islands were removed by 26-connected component analysis, and cavities were removed by 6-connected component analysis, yielding a fully connected blood vessel segment (**D5**). A different threshold was chosen to derive the tubule segment (**D6**). As the tubule segment contained discontinuities in regions where tubules were compressed by nearby distended large blood vessels, connected component analysis could not be applied.

Maximum inscribed ball

We extended the thickness measurement algorithm proposed by Hildebrand and Ruegsegger[13], based on subpixel-accurate signed distance transformation, in order to take advantage of the aggregate compute power of data centers with Message Passing Interface (MPI).

Manual annotations for glomeruli and perirenal fat

To generate manual training data for glomeruli, three selected regions of interest (ROI) 512 × 256 × 256 voxels in size were segmented by manual contouring, using the Freehand Selection Tool in Fiji/ImageJ.[14] The resulting binary masks were viewed and corrected in all three dimensions multiple times in order to minimize slice-by-slice discontinuities common to single-dimension contouring ("stacks of pancakes"). Residual discontinuities remain in the last manually contoured dimension, which could in principle be removed by smoothing, but were kept as is in order to keep the annotations fully manual (**D4**).

For removing perirenal fat from the blood vessel segment for visualization, three different regions of interest were annotated. Manual work was reduced by providing an automated initial guess of what was believed to be fat. For this, the image was strongly low-pass filtered and thresholded. The thickness transform of the resulting binary mask was calculated and thresholded to provide the initial guess, which was then corrected by manual annotation (**D9**).

Scattering transform

In order to detect glomeruli and remove perirenal fat from the blood vessel segment, we used a ML-based approach based on invariant scattering convolution networks.[15] The scattering transform allowed us to design networks producing feature maps that are stable under small deformations. Although these networks are not steerable in the strictest sense, since they are not relying on irreducible representations of SO3, the produced features are equivariant under the action of a discrete subgroup of SO3 (regular representation).

The manually annotated regions of interest represent 0.06% of the image. The training data were supplemented by additional regions of interest which contained no glomeruli or no fat at all, and thus did not require manual annotation. The supervised set was then used to train a hybrid algorithm that relied on a 3D scattering transform convolutional network topped with a dense neural network. Gray values of the images were convolved with kernels with specific band pass properties and orientation, then contracted using complex modulus to generate intermediate scalar features. These features were further convolved with a low pass filter kernel to enforce the emergence of local invariances. The generated feature maps were fed to a dense, feed-forward neural network. The neural network processes each pixel independently to come up with a Boolean answer for each pixel. The neural network inference was run everywhere before further processing the foreground.

The scattering transform relied upon ad-hoc designed 3D kernels (Morlet's wavelet with different sizes and orientations) that uniformly covered all directions at different scales, following the original work on the invariant scattering convolution networks.[15] In the scattering convolutional network, filter nonlinearities were obtained by taking the magnitude of the filter responses and convolving them again with the kernels in a cascading fashion. These nonlinearities are designed to be robust against small Lipschitz-continuous deformations of the image.[15]

As opposed to our curvelet-based image enhancement approach, we decomposed the image into image cubic tiles, then applied a windowed -- thus local -- Fourier transform on a tile by considering a region about twice as large by width around the tile. While it would be possible to use a convolutional network based upon a global scattering transform, it would have incurred a very large amount of features that were to be consumed at once, leading to an intermediate footprint of about one Petabyte that would exceed the available RAM capacity of the cluster.

The scattering transform convolutional network produced a stack of a few hundreds of scalar feature maps, per pixel. We would like to note that if considered as a "fiber bundle"[16,17], the feature map stack is equivariant under the symmetry group of rotations (i.e. the stack is a regular representation of SO(3)). This property can be exploited by further processing the feature maps with a dense neural network with increased parameter sharing across the hidden layers, making the output layer invariant to rotations.

The foreground representing the glomeruli was followed by a connected component analysis; the components are further examined by a tiny support vector machine to decide whether to retain or eliminate the connected component under examination (vector dimensionality of 8, training set of about 50 samples).[18] The SVM feature vector included mass thickness percentiles (1, 10, 20, 40, 60, 80, 90, 99), fractal dimension and moment of inertia. Glomerular volume can be expected to lie within a restricted range.

<u>Validation of glomeruli by domain expert</u>
To reduce the workload to the level required to make validation of each glomerulus feasible, a volumetric visualization of the overlap between the blood vessel mask and the glomerular mask was generated for rapid evaluation of easy to recognize glomeruli or artefacts. In a first round, glomeruli were classified by their shape into three categories: 1. certain true positive with shape distortion, 2. certain true positive without shape distortion and 3. uncertain. Candidates of the third category, which constituted about 3 – 5 % of all machine-identified glomeruli, were then reviewed in a second round as an overlay over the original raw data on a slice-by-slice basis. These candidates were then assigned as false positives or as certain glomeruli of categories 1 or 2.